%% file: main.tex
\documentclass{article}

\usepackage{microtype}
\usepackage{graphicx}
\usepackage{subcaption}
\usepackage{booktabs} %

\usepackage{hyperref}

 \usepackage[accepted]{icml2026}

\usepackage{amsmath}
\usepackage{amssymb}
\usepackage{mathtools}
\usepackage{amsthm}
\usepackage{url}
\usepackage{caption}  
\usepackage{multirow}
\usepackage{makecell} %
\usepackage{amsmath}  
\usepackage{bm}
\usepackage{xcolor}
\usepackage{hyperref}

\usepackage[capitalize,noabbrev]{cleveref}

\theoremstyle{plain}

\theoremstyle{definition}

\theoremstyle{remark}

\usepackage[disable,textsize=tiny]{todonotes}

\icmltitlerunning{ZeroDiff: Zero-Shot Time Series Reconstruction via Informed-Prior Diffusion}

\begin{document}

\twocolumn[
\icmltitle{ZeroDiff: Zero-Shot Time Series Reconstruction via Informed-Prior Diffusion}

\begin{icmlauthorlist}
\icmlauthor{Yingda Fan}{pitt}
\icmlauthor{Dan Lu}{ornl}
\icmlauthor{Xiaowei Jia}{pitt}
\end{icmlauthorlist}

\icmlaffiliation{pitt}{Department of Computer Science, University of Pittsburgh, Pittsburgh, PA, USA}
\icmlaffiliation{ornl}{Oak Ridge National Laboratory, Oak Ridge, TN, USA}

\icmlcorrespondingauthor{Xiaowei Jia}{xiaowei@pitt.edu}

\icmlkeywords{Time Series Reconstruction, Diffusion Model, Zero-Shot Learning, Informed Prior}

\vskip 0.3in
]

\printAffiliationsAndNotice{}

\urlstyle{same}  %

\begin{abstract}
Time series modeling increasingly demands high-quality supervision, yet target observations remain scarce—exogenous inputs are broadly available, but target measurements are often unavailable due to cost, infrastructure, or accessibility constraints. Can models trained on observed locations reconstruct target time series where measurements have never been collected? We term this \textit{zero-shot time series reconstruction}. A naive approach—directly mapping exogenous inputs to targets—can yield predictions at unobserved locations, but without target signals, such models fail to capture the intrinsic dynamics of the target variable, producing overly smooth outputs that underestimate extremes. This reveals systematic errors that call for explicit modeling and calibration. We propose \textbf{ZeroDiff}, which constructs an informed prior from exogenous variables alone, then learns to calibrate reconstruction errors through diffusion—training on observed locations and generalizing to unobserved ones. Experiments across diverse real-world datasets demonstrate significant improvements over existing approaches. Our code is available at %
\url{https://github.com/YingdaFan/ZeroDiff-ICML2026}.

\end{abstract}

\input{1-Introduction}

 \input{2-RelatedWork}

\input{3-Preliminary}

 \input{4-Method}

 \input{5-Experiment}

\input{6-Conclusion}

\newpage

\section*{Acknowledgments}
This research is supported by Dan Lu's Early Career Project, sponsored by the Office of Biological and
Environmental Research in the U.S. Department of Energy (DOE). 
X.J. was partially supported by the National Science Foundation (NSF) grants 2203581, 2239175, 2316305, 2147195, 2425845, and 2530609; the USGS award  G22AC00266; and the NASA grants 80NSSC24K1061 and 80NSSC25K0013.

\section*{Impact Statement}
This paper presents ZeroDiff for zero-shot time series reconstruction, with direct applications in hydrological modeling. Potential positive societal impacts include improved flood early warning, water resource management, and environmental monitoring in ungauged regions where monitoring infrastructure is limited. As with any predictive model, we encourage practitioners to appropriately communicate uncertainty estimates when informing real-world decisions.

\bibliography{main}
\bibliographystyle{icml2026}

\appendix

\onecolumn
\input{7-Appendix}

\end{document}

%% file: 1-Introduction.tex
\section{Introduction}
As time series modeling scales from specialized architectures~\cite{nie2023patchtst, liu2024itransformer} to foundation models~\cite{ansari2024chronos, das2024timesfm}, the need for high-quality supervision has grown—yet in many scenarios, target observations remain scarce or entirely missing. 
This challenge arises across domains—from flood forecasting~\cite{kratzert2018rainfall}, where meteorological data are abundant but streamflow is sparsely observed across space, to solar energy prediction~\cite{paletta2023advances}, where satellite imagery and reanalysis data are globally available but  irradiance observations are measured at few stations. In each case, exogenous inputs are broadly available, but target measurements are observed only at a subset of locations. In such settings, practitioners face a fundamental question: can models learn from locations where targets are observed to reconstruct target time series at locations where they have never been measured? We term this problem \textit{zero-shot cross-domain time series reconstruction}. Unlike forecasting~\cite{zhou2021informer, salinas2020deepar}, which predicts future values from historical observations at the same location, 
or imputation~\cite{cao2018brits, du2023saits}, which fills in missing values given partial observations, zero-shot reconstruction requires the model to infer an entire target time series at locations with no historical target data whatsoever. This setting poses a distinct challenge: the model must generalize across locations, reconstructing temporal dynamics it has never directly observed.

A central difficulty lies in the heterogeneity of target distributions 
across locations. While the temporal dynamics—how targets respond to 
exogenous inputs—are often governed by shared physical or statistical 
principles, the marginal distribution of the target variable can vary 
dramatically in mean and variance. As a result, models trained on 
observed locations struggle to transfer to unobserved 
ones.

In such settings, probabilistic reconstruction is essential. When ground truth targets are unavailable, uncertainty quantification becomes important for downstream decision-making~\cite{gneiting2014probabilistic}.  Diffusion models~\cite{ho2020denoising, song2021scorebased} offer a principled framework for probabilistic modeling~\cite{rasul2021autoregressive, tashiro2021csdi}, but standard diffusion requires access to $\mathbf{Y}_0$ for both forward corruption and loss computation—impossible when targets are absent. Recent work shows diffusion benefits from informative initialization: CARD~\cite{han2022card} replaces $\mathcal{N}(\mathbf{0}, \mathbf{I})$ with a regression-based prior, SDEdit~\cite{meng2022sdedit} preserves structure via partial noising. These advances suggest that if we can construct an informed prior $\hat{\mathbf{Y}}$ capturing the expected target, diffusion can shift from generation to calibration—refining a structured estimate rather than generating from noise.

\begin{figure}[!th]
\vskip 0.05in
\centering
\includegraphics[width=\columnwidth]{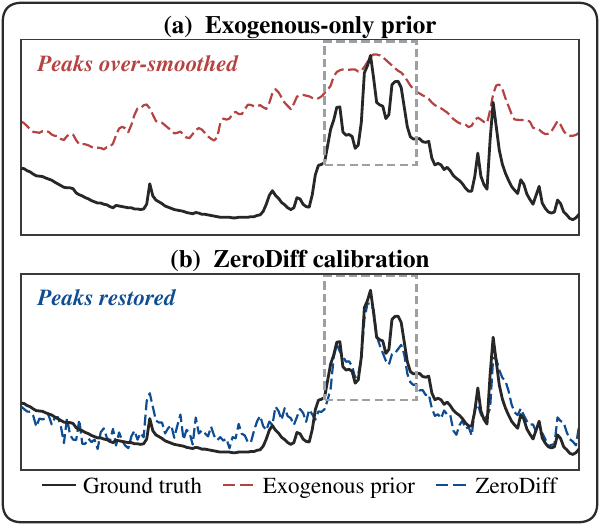}
\caption{In the exogenous-only reconstruction the peaks are over-smoothed and no longer stand out \textbf{(a)}; ZeroDiff calibrates it onto the ground truth, restoring their prominence \textbf{(b)}.}
\label{fig:introduction}
\vskip -0.15in 
\end{figure}

We present \textbf{ZeroDiff}, a diffusion-based framework for 
zero-shot time series reconstruction. It decomposes the 
cross-location generalization problem into two stages connected 
by a shared informed prior $\hat{\mathbf{Y}}$. The first stage generalizes from 
exogenous inputs to target structure: target distributions vary 
drastically across locations in mean and variance, but their 
input–output dynamics are shared once expressed in a standardized 
space. The location-specific statistics needed for that 
standardization can themselves be inferred from exogenous variables 
alone. The informed prior $\hat{\mathbf{Y}}$ at unobserved 
locations is then constructed directly from $\mathbf{X}$, without 
any target signal.

The second stage generalizes calibration across locations. Because 
$\hat{\mathbf{Y}}$ is constructed identically at observed and 
unobserved locations, its systematic errors share the same structure 
everywhere. A diffusion model trained to calibrate the prior on 
observed locations therefore transfers to unobserved ones. This 
shifts the role of diffusion from generation to calibration: the 
reverse process refines $\hat{\mathbf{Y}}$ rather than denoising from 
$\mathcal{N}(\mathbf{0}, \mathbf{I})$.

In summary, our contributions are:
\begin{itemize}
    \item We formalize the problem of \textbf{zero-shot time series 
    reconstruction} and identify a structural assumption that makes 
    it tractable: target magnitudes vary across locations, but 
    input–target dynamics are shared.
    
    \item We propose \textbf{informed-prior diffusion}: a single 
    prior inferred from exogenous variables both carries 
    cross-location dynamics and anchors a calibration that transfers 
    from observed to unobserved locations.
    
    \item We design a bidirectional denoiser that conditions on both past and future timesteps, improving reconstruction at peaks and sharp transitions.
\end{itemize}

\paragraph{Conflict of Interest Disclosure.}
The authors declare no financial conflicts of interest related to this work.

%% file: 2-RelatedWork.tex
\section{Related Work}

\textbf{Diffusion Models for Time Series.}
Diffusion models have recently emerged as a powerful framework for probabilistic time series modeling, achieving strong performance across forecasting, imputation, and generation tasks. TimeGrad~\cite{rasul2021autoregressive} first introduces autoregressive denoising diffusion for probabilistic forecasting, generating future values step-by-step conditioned on historical observations. CSDI~\cite{tashiro2021csdi} extends score-based diffusion to imputation via conditional generation. SSSD~\cite{alcaraz2022diffusion} integrates structured state-space models to capture long-range dependencies. TimeDiff~\cite{shen2023timediff} proposes non-autoregressive diffusion for parallel generation. Diffusion-TS~\cite{yuan2024diffusionts} employs encoder-decoder transformers with disentangled temporal representations. mr-Diff~\cite{shen2024mrdiff} leverages multi-resolution structure through seasonal-trend decomposition. Most recently, CNDiff~\cite{cndiff2025} introduces nonlinear data transformations for conditional forecasting. Despite these advances, all existing diffusion-based time series methods share a fundamental assumption: target observations must be available at locations during training. 
This assumption is embedded at multiple levels—normalization, forward corruption, and loss computation—all of which require access to ground-truth targets. 
Consequently, when targets are entirely absent at a subset of locations, the diffusion framework cannot be applied. Our work addresses this previously unsupported regime.

\textbf{Diffusion with Informative Priors.}
Standard diffusion models assume the forward process terminates at an uninformative prior $\mathcal{N}(\mathbf{0}, \mathbf{I})$~\cite{ho2020denoising, song2021scorebased}, requiring the reverse process to generate from scratch. Recent work in computer vision has challenged this assumption by incorporating \emph{data-dependent priors}. PriorGrad~\cite{lee2021priorgrad} demonstrates that adaptive priors derived from conditional information improve both efficiency and sample quality. CARD~\cite{han2022card} replaces the standard endpoint with a regression-based prior $\mathcal{N}(f(\mathbf{x}), \sigma)$, where $f(\mathbf{x})$ is a pre-trained predictor; the diffusion process then \emph{refines} predictions rather than generates from noise. SDEdit~\cite{meng2022sdedit} starts the reverse process from partially noised inputs, preserving structural information from a guide image. Cold Diffusion~\cite{bansal2023cold} demonstrates that meaningful degradations—such as blur or downsampling—can replace Gaussian noise entirely, revealing that diffusion's core mechanism is learning to invert transformations. 
These approaches share a common insight: diffusion is most effective as a refinement mechanism when initialized from a structured prior, a design principle that has also influenced other prior-guided reconstruction pipelines~\cite{yuan2025dvp, yuan2025msp}.
However, existing methods assume that the prior itself is obtained from supervised learning using target observations at the same domain. In contrast, our setting requires constructing informative priors without any target data at the reconstruction location. \textbf{ZeroDiff} extends the idea of informative priors to a zero-shot regime, where priors are inferred solely from exogenous variables and calibrated via diffusion using supervision from other locations.

\textbf{Zero-shot Learning and Cross-location Generalization.}
Zero-shot learning enables recognition of classes unseen during training by leveraging auxiliary information~\cite{lampert2009learning}. Early approaches learn mappings between visual features and semantic attributes~\cite{farhadi2009describing} or word embeddings~\cite{frome2013devise}. Subsequent work explores compatibility functions~\cite{akata2016label}, latent embeddings~\cite{xian2016latent}, and generative approaches~\cite{xian2018feature}. Xian et al.~\cite{xian2017zero} provide a comprehensive benchmark establishing evaluation protocols for the field. The core principle—transferring knowledge from seen to unseen classes via shared semantic representations—motivates our approach of transferring dynamics from observed to unobserved locations.

In spatio-temporal modeling, cross-location generalization has been explored through graph neural networks. STGCN~\cite{yu2018spatio} and DCRNN~\cite{li2018diffusion} capture spatial dependencies via graph convolutions. Recent work addresses cross-city transfer: Domain Adversarial Spatial-Temporal Networks~\cite{tang2022domain} learn city-invariant representations, while Graph Neural Processes~\cite{hu2023graph} enable extrapolation to unobserved locations.  However, these methods assume that at least some target observations are available at all locations, focusing on forecasting rather than reconstruction. They do not address the setting where targets are unobserved at test locations.

Time series foundation models such as Chronos~\cite{ansari2024chronos}, TimesFM~\cite{das2024timesfm}, and Time-MoE~\cite{shi2025timemoe} demonstrate impressive zero-shot capabilities across datasets. However, their ``zero-shot'' refers to generalization across \emph{datasets}—not across locations within a dataset where some locations lack target 
observations entirely. 

In hydrology, the problem of Prediction in Ungauged Basins (PUB)~\cite{kratzert2019toward} is conceptually related, aiming to predict streamflow at locations without gauges. While recent machine learning approaches improve point estimates in this setting,  they typically lack principled uncertainty quantification.

%% file: 3-Preliminary.tex
\section{Preliminary}

\subsection{Problem Formulation}

\paragraph{Zero-shot time series reconstruction.} Given exogenous variables $\mathbf{X}$ at a new location, the goal is to reconstruct the target time series $\mathbf{Y}$ without any historical target observations at that location.

Formally, consider $K$ locations, each with exogenous variables $\mathbf{X}^{(k)} \in \mathbb{R}^{L \times D}$ and target time series $\mathbf{Y}^{(k)} \in \mathbb{R}^{L}$, where $\mathbf{X}$ and $\mathbf{Y}$ are temporally aligned with sequence length $L$~\cite{tashiro2021csdi, cao2018brits}. Locations are partitioned into:
\begin{itemize}
    \item \textbf{Observed} $\mathcal{O}$: both $\mathbf{X}^{(k)}$ and $\mathbf{Y}^{(k)}$ are available for training.
    \item \textbf{Unobserved} $\mathcal{U}$: only $\mathbf{X}^{(j)}$ is accessible; the goal is to reconstruct $\mathbf{Y}^{(j)}$.
\end{itemize}

Both training and inference operate on the same temporal span. This differs from forecasting~\cite{zhou2021informer, wu2021autoformer, nie2023patchtst, liu2024itransformer}, which predicts future from past at the \emph{same} location, and from imputation~\cite{tashiro2021csdi, cao2018brits, du2023saits}, which fills gaps given \emph{partial} observations. Here, the model must generalize across locations to reconstruct targets it has never seen.

The central challenge is that while the input-target dynamics may be shared across locations once properly standardized, the marginal distribution of $\mathbf{Y}$ differs substantially in mean and variance across locations. Standard normalization techniques~\cite{kim2021reversible, liu2022non} require target observations to compute location-specific statistics $(\mu, \sigma)$---precisely what is unavailable for $j \in \mathcal{U}$.

\subsection{Diffusion Models for Time Series}

Denoising diffusion probabilistic models (DDPMs)~\cite{ho2020denoising, song2021scorebased, luo2022understandingdiffusion} learn data distributions through a forward process that progressively adds noise and a reverse process that learns to denoise. Given target $\mathbf{Y}_0$, the forward process is defined as:
\begin{equation}
    q(\mathbf{Y}_t|\mathbf{Y}_0) = \mathcal{N}\bigl(\sqrt{\bar\alpha_t}\mathbf{Y}_0, (1-\bar\alpha_t)\mathbf{I}\bigr)
\end{equation}
The reverse process learns to denoise by optimizing $\mathcal{L}_{\text{DDPM}} = \mathbb{E}_{t, \mathbf{Y}_0, \boldsymbol{\epsilon}} \bigl[ \|\boldsymbol{\epsilon} - \boldsymbol{\epsilon}_\theta(\mathbf{Y}_t, t)\|^2 \bigr]$.\label{eq:ddpm_loss}
For time series applications, two formulations are common depending on whether temporal windows are employed:

\textbf{Sliding window formulation.} In forecasting tasks~\cite{rasul2021autoregressive, li2024tmdm, shen2024multi, yuan2024diffusionts, gao2025armd}, the model conditions on historical observations $\mathbf{X} \in \mathbb{R}^{N \times D}$ to predict future targets $\mathbf{Y} \in \mathbb{R}^{M \times D}$, where $N$ and $M$ denote the lookback and prediction horizon lengths respectively. 

\textbf{Aligned formulation.} In reconstruction or imputation tasks~\cite{tashiro2021csdi, alcaraz2022diffusionimputation, kollovieh2023predict}, $\mathbf{X} \in \mathbb{R}^{L \times D}$ and $\mathbf{Y} \in \mathbb{R}^{L}$ share the same temporal span $L$. The model learns to recover $\mathbf{Y}$ from temporally aligned exogenous inputs.

A key advancement is the use of \emph{informed priors}~\cite{han2022card, meng2022sdedit, li2024tmdm}. Rather than diffusing toward an uninformative endpoint $\mathcal{N}(\mathbf{0}, \mathbf{I})$, these methods use a regression-based prior $\mathcal{N}(\hat{\mathbf{Y}}, \sigma^2\mathbf{I})$, where $\hat{\mathbf{Y}} = f(\mathbf{X})$. The forward process becomes:
\begin{equation}
\label{eq:informed_forward_standard}
    q(\mathbf{Y}_t|\mathbf{Y}_0, \mathbf{X}) = \mathcal{N}\bigl(\sqrt{\bar\alpha_t}\mathbf{Y}_0 + (1-\sqrt{\bar\alpha_t})\hat{\mathbf{Y}}, \bar\sigma_t \mathbf{I}\bigr)
\end{equation}
where $\bar\sigma_t = 1 - \bar\alpha_t$.

\textbf{The missing-target problem.} Crucially, both formulations require access to ground-truth targets $\mathbf{Y}_0$ during training. The forward process in Eq.~\eqref{eq:informed_forward_standard} constructs $\mathbf{Y}_t$ from $\mathbf{Y}_0$, and the denoising process computes the loss using $\mathbf{Y}_t$. For unobserved locations, neither the forward process nor the training objective can be defined—the diffusion framework fundamentally breaks down.

%% file: 4-Method.tex
\section{Methodology}

\begin{figure*}[!t]
    \centering
    \includegraphics[width=\linewidth]{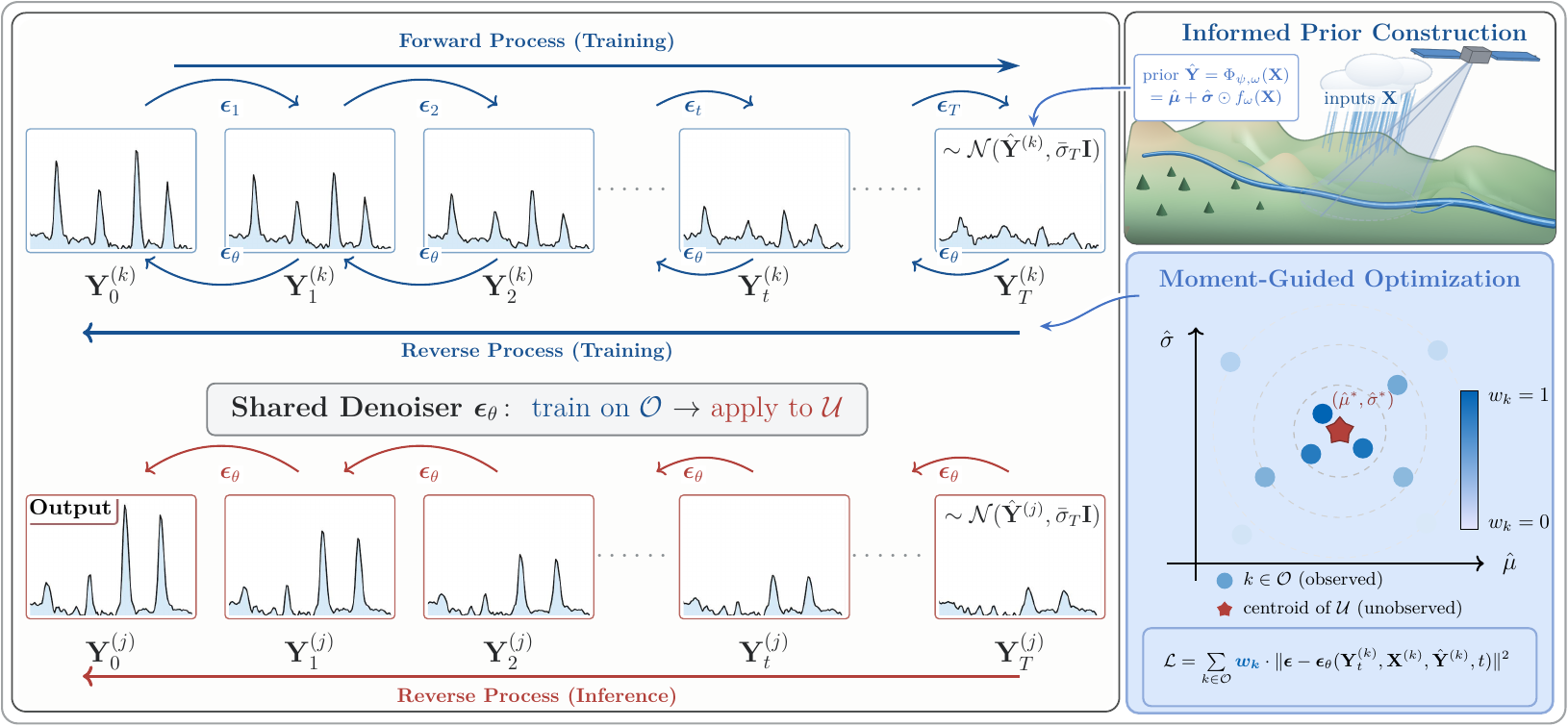}
    \caption{Proposed framework. The informed prior $\hat{\mathbf{Y}} = \Phi_{\psi,\omega}(\mathbf{X})$ combines moment estimation (VAE) with dynamics learning in standardized space, generalizing to unobserved locations. The diffusion process starts from $\mathcal{N}(\hat{\mathbf{Y}}, \bar\sigma_T \mathbf{I})$ rather than pure noise, enabling calibration instead of generation. The estimated moments also guide optimization by weighting training locations based on their proximity to the target in moment space.}

    \label{fig:framework}
    \vspace{-0.1in}
\end{figure*}

We propose a framework for zero-shot time series reconstruction, illustrated in Fig.~\ref{fig:framework}, built on three insights:
(i) while target marginals $p(\mathbf{Y})$ vary drastically across locations in mean and standard deviation, the input-target dynamics are transferable once standardized; 
(ii) location-specific statistics $(\mu, \sigma)$ can be inferred from exogenous variables $\mathbf{X}$ alone via cross-modal learning, without requiring target observations; 
(iii) by constructing informed priors for $\mathcal{O}$ where ground-truth is available, the diffusion model learns to calibrate estimation errors on $\mathcal{O}$, and this calibration generalizes to $\mathcal{U}$.
{Motivated by these observations, we first construct an informed prior through moment estimation and dynamics learning, and then refine it through diffusion-based calibration}.

\subsection{Transferable Prior Construction}%
\label{sec:moment_dynamics}

Recent work on non-stationary time series~\citep{kim2021reversible, liu2022non} demonstrates that separating statistics (mean, variance) from temporal dynamics improves generalization. While the marginal distribution of $\mathbf{Y}$ varies across locations in magnitude, the temporal dynamics of how $\mathbf{X}$ drives $\mathbf{Y}$, once normalized, are shared. This motivates learning in normalized space, but standard normalization requires target observations to compute location-specific $(\mu, \sigma)$, which is precisely unavailable for $j \in \mathcal{U}$.

We address this limitation through \emph{cross-modal moment estimation}: inferring target statistics $(\mu_k, \sigma_k)$ from exogenous variables $\mathbf{X}^{(k)}$ alone. Although $\mathbf{X}$ and $\mathbf{Y}$ represent different physical quantities, the distributional characteristics of $\mathbf{X}$ are predictive of the magnitude of $\mathbf{Y}$. In hydrological systems, for instance, a basin's 
weather data and catchment properties can affect the magnitude of runoff. %

Specifically, we model the marginal moments of $\mathbf{Y}^{(k)}$ as a learnable function of exogenous summary statistics:
\begin{equation}
\label{eq:moment_mapping}
    (\mu_k, \sigma_k) = h(\mathbf{Z}^{(k)})
\end{equation}
where $\mathbf{Z}^{(k)} = [\bar{\mathbf{X}}^{(k)}; \sigma_{\mathbf{X}}^{(k)}] \in \mathbb{R}^{2D}$ concatenates the temporal mean and standard deviation of exogenous inputs.
Since $(\mu_k, \sigma_k)$ are time-invariant statistics, we condition on summary statistics of $\mathbf{X}$ rather than the full sequence.

\paragraph{Cross-modal moment estimation.} 
We estimate the target moments using a conditional VAE~\citep{kingma2014auto, sohn2015learning}—a form of exogenous-conditioned prior estimation that captures distributional structure across locations.
The architecture consists of three components: a \emph{latent encoder} $q_\psi(\mathbf{z}|\mu_k, \sigma_k)$ that maps target moments to a latent distribution, capturing location-specific deviations from the mean behavior; a \emph{feature encoder} $\phi(\mathbf{Z}^{(k)})$ that transforms exogenous statistics into a representation conditioning the decoder; and a \emph{decoder} $p_\psi(\mu_k, \sigma_k|\mathbf{z}, \phi(\mathbf{Z}^{(k)}))$ that reconstructs target moments from both the latent code and the encoded exogenous features.

During training on $\mathcal{O}$, we optimize the evidence lower bound:
\begin{equation}
\label{eq:vae_loss}
\begin{split}
    \mathcal{L}_{\text{VAE}} &= \mathbb{E}_{q_\psi}\left[\|(\mu_k, \sigma_k) - \mathrm{Dec}_\psi(\mathbf{z}, \phi(\mathbf{Z}^{(k)}))\|^2\right] \\
    &\quad + \lambda \cdot D_{\mathrm{KL}}\bigl(q_\psi(\mathbf{z}|\mu_k, \sigma_k) \,\|\, \mathcal{N}(\mathbf{0}, \mathbf{I})\bigr)
\end{split}
\end{equation}
The KL regularization encourages the latent space to capture shared distributional structure: locations with similar exogenous characteristics are mapped to nearby points, enabling moment estimation to generalize from $\mathcal{O}$ to $\mathcal{U}$.

For $j \in \mathcal{U}$, although $\mathbf{Y}^{(j)}$ is unavailable, we can compute $\mathbf{Z}^{(j)} = [\bar{\mathbf{X}}^{(j)}; \mathrm{std}(\mathbf{X}^{(j)})]$ directly from $\mathbf{X}^{(j)}$. We then decode using the prior mean:
\begin{equation}
\label{eq:vae_inference}
    (\hat{\mu}_j, \hat{\sigma}_j) = \mathrm{Dec}_\psi(\mathbf{0}, \phi(\mathbf{Z}^{(j)}))
\end{equation}
This enables zero-shot moment estimation: using only $\mathbf{X}^{(j)}$ and the distributional structure learned from $\mathcal{O}$, we infer $(\hat{\mu}_j, \hat{\sigma}_j)$ without any target observations.

\paragraph{Dynamics learning in standardized space}
We learn a \emph{single} dynamics model $f_\omega: \mathbb{R}^{L \times D} \to \mathbb{R}^{L}$ shared across all observed locations, performing reconstruction over the same temporal span rather than forecasting. 
The dynamics model $f_\omega$ can adopt architectures such as LSTM~\citep{kratzert2018rainfall}, DLinear~\citep{zeng2023transformers}, Mamba~\citep{gu2023mamba}, or TimesNet~\citep{wu2023timesnet}; we compare these choices in Table~\ref{tab:dynamic_models}.

The training objective aggregates prediction error over $\mathcal{O}$:
\begin{equation}
\label{eq:lstm_training}
    \min_\omega \sum_{k \in \mathcal{O}} 
    \left\| f_\omega(\mathbf{X}^{(k)}) - \tilde{\mathbf{Y}}^{(k)} \right\|^2
\end{equation}
where $\tilde{\mathbf{Y}}^{(k)} = (\mathbf{Y}^{(k)} - \mu_k)/\sigma_k$ is the standardized target. By sharing parameters and training in normalized space, $f_\omega$ captures the shared temporal dynamics of how $\mathbf{X}$ drives fluctuations in $\mathbf{Y}$, rather than fitting location-specific magnitudes.

\paragraph{Informed prior construction.}
We use VAE-estimated statistics $(\hat{\mu}, \hat{\sigma})$ for both $\mathcal{O}$ and $\mathcal{U}$—even for $k \in \mathcal{O}$ where ground-truth statistics $(\mu_k, \sigma_k)$ are available. This ensures consistent estimation of error characteristics between training and inference, enabling the downstream diffusion model to learn a calibration that generalizes to $\mathcal{U}$.
The informed prior for $k \in \mathcal{O}$:
\begin{equation}
\label{eq:prior_train}
    \hat{\mathbf{Y}}^{(k)} = \hat{\mu}_k + \hat{\sigma}_k \cdot f_\omega(\mathbf{X}^{(k)})
\end{equation}
For $j \in \mathcal{U}$:
\begin{equation}
\label{eq:prior_test}
    \hat{\mathbf{Y}}^{(j)} = \hat{\mu}_j + \hat{\sigma}_j \cdot f_\omega(\mathbf{X}^{(j)})
\end{equation}
where $(\hat{\mu}_j, \hat{\sigma}_j)$ is obtained via Eq.~\eqref{eq:vae_inference}. For unobserved locations, the estimated moments play a critical role: $f_\omega$ captures the normalized response pattern, while $(\hat{\mu}_j, \hat{\sigma}_j)$ rescales it to the appropriate magnitude. For notational convenience, we denote the composite mapping as $\Phi_{\psi,\omega}(\mathbf{X}) := \hat{\mu} + \hat{\sigma} \cdot f_\omega(\mathbf{X})$, where $(\hat{\mu}, \hat{\sigma}) = \mathrm{Dec}_\psi(\mathbf{0}, \phi(\mathbf{Z}))$.

\subsection{Generalizable Diffusion Calibration}
The informed prior $\hat{\mathbf{Y}}^{(k)}$ and $\hat{\mathbf{Y}}^{(j)}$ provide an initial reconstruction but inherit estimation error from both moment estimation and dynamics learning. We refine this prior through diffusion, learning the conditional distribution $p(\mathbf{Y}^{(k)} | \hat{\mathbf{Y}}^{(k)}, \mathbf{X}^{(k)})$ on $\mathcal{O}$. The denoiser $\boldsymbol{\epsilon}_\theta$ is shared across all $k \in \mathcal{O}$, learning a calibration that generalizes to $\mathcal{U}$. At inference, we apply this learned distribution to $j \in \mathcal{U}$ by conditioning on $(\hat{\mathbf{Y}}^{(j)}, \mathbf{X}^{(j)})$. Since the estimation procedure is shared, the error characteristics remain consistent across $\mathcal{O}$ and $\mathcal{U}$, enabling the calibration to generalize.

\paragraph{Informed-prior forward process.} We diffuse toward the informed prior rather than zero:
\begin{equation}
\label{eq:informed_forward}
    q(\mathbf{Y}_t^{(k)}|\mathbf{Y}^{(k)}, \mathbf{X}^{(k)}) = \mathcal{N}\bigl(\sqrt{\bar\alpha_t}\mathbf{Y}^{(k)} + (1-\sqrt{\bar\alpha_t})\hat{\mathbf{Y}}^{(k)}, \bar\sigma_t \mathbf{I}\bigr)
\end{equation}
The noise schedule $(\bar\alpha_t, \bar\sigma_t)$ is shared across all locations, while each location $k$ has its own informed prior $\hat{\mathbf{Y}}^{(k)} = \hat{\mu}_k + \hat{\sigma}_k \cdot f_\omega(\mathbf{X}^{(k)})$ as defined in Eq.~\eqref{eq:prior_train}, incorporating $\mathbf{X}^{(k)}$ through both the VAE-estimated moments and the dynamics model. 
At $t=T$, this yields the endpoint distribution $\mathbf{Y}_T^{(k)} \sim \mathcal{N}(\hat{\mathbf{Y}}^{(k)}, \bar\sigma_T \mathbf{I})$, in contrast to the standard $\mathcal{N}(\mathbf{0}, \mathbf{I})$ used in conventional diffusion models. This formulation follows the informed-prior framework of~\citet{han2022card}.

\paragraph{Prior-anchored reverse process.} 
The informed-prior forward process (Eq.~\eqref{eq:informed_forward}) induces a modified posterior distribution. At each reverse step, we sample from $p_\theta(\mathbf{Y}_{t-1}^{(k)} | \mathbf{Y}_t^{(k)}, \mathbf{X}^{(k)}, \hat{\mathbf{Y}}^{(k)})$, whose posterior mean takes the form:
\begin{equation}
\label{eq:posterior_mean}
    \boldsymbol{\mu}_{t-1}^{(k)} = \gamma_0 \mathbf{Y}^{(k)} + \gamma_1 \mathbf{Y}_t^{(k)} + \gamma_2 \hat{\mathbf{Y}}^{(k)}
\end{equation}
where $\gamma_0, \gamma_1, \gamma_2$ are functions of $\alpha_t, \bar{\alpha}_{t-1}$, and $\bar{\sigma}_t$ (see Appendix~\ref{appendix:posterior_derivation}). Unlike standard DDPM, our posterior includes a third term—anchoring the trajectory to $\hat{\mathbf{Y}}$.

During inference for $j \in \mathcal{U}$, the clean target $\mathbf{Y}^{(j)}$ is unavailable. We estimate it via reparameterization:
\begin{equation}
\label{eq:reparam}
    \hat{\mathbf{Y}}_0^{(j)} = \frac{1}{\sqrt{\bar{\alpha}_t}}\bigl(\mathbf{Y}_t^{(j)} - (1 - \sqrt{\bar{\alpha}_t})\hat{\mathbf{Y}}^{(j)} - \sqrt{\bar{\sigma}_t}\boldsymbol{\epsilon}_\theta\bigr)
\end{equation}
Substituting $\hat{\mathbf{Y}}_0^{(j)}$ for $\mathbf{Y}^{(j)}$ in Eq.~\eqref{eq:posterior_mean} yields $\boldsymbol{\mu}_{t-1}^{(j)}$, ensuring the reverse trajectory remains guided by the informed prior throughout denoising.

\paragraph{Moment-guided optimization.} 
In addition to denormalization in Eq.~\eqref{eq:prior_test}, the estimated moments $(\hat{\mu}_k, \hat{\sigma}_k)$ guide the optimization of the shared denoiser $\boldsymbol{\epsilon}_\theta$. 
Since every $k \in \mathcal{O}$ contributes to updating the shared denoiser parameters, we reweight each location's gradient contribution based on its proximity to the target in moment space.
Let $(\hat{\mu}^*, \hat{\sigma}^*)$ denote the estimated moments of the target location; for simultaneous reconstruction of multiple targets, we use the centroid (Appendix~\ref{appendix:multi-target}).
We define weights using a Gaussian kernel over the moment space:
\begin{equation}
\label{eq:weights}
    w_k = \exp\left(-\frac{\|(\hat{\mu}_k, \hat{\sigma}_k) - (\hat{\mu}^*, \hat{\sigma}^*)\|^2}{2\tau^2}\right)
\end{equation}
where $\tau$ controls the bandwidth. The training objective becomes:
\begin{equation}
\label{eq:loss}
    \mathcal{L} = \sum_{k \in \mathcal{O}} w_k \cdot \mathbb{E}_{t, \boldsymbol{\epsilon}} \left[ \|\boldsymbol{\epsilon} - \boldsymbol{\epsilon}_\theta(\mathbf{Y}_t^{(k)}, \mathbf{X}^{(k)}, \hat{\mathbf{Y}}^{(k)}, t)\|^2 \right]
\end{equation}
Rather than aggregating gradients equally across all $k \in \mathcal{O}$, this formulation upweights locations whose  moments are close to $(\hat{\mu}^*, \hat{\sigma}^*)$, yielding a denoiser better suited for $j \in \mathcal{U}$.

\paragraph{Bidirectional denoiser.} 
Our calibration setting 
is fundamentally different from prior 
diffusion-based forecasting methods %
~\citep{rasul2021autoregressive}, which adopts causal masking to prevent information leakage from future to past~\citep{tashiro2021csdi, shen2023timediff}. %
In contrast, 
during the  calibration process $p(\mathbf{Y}^{(k)} | \hat{\mathbf{Y}}^{(k)}, \mathbf{X}^{(k)})$, the conditioning signals $\mathbf{X}^{(k)}$ and $\hat{\mathbf{Y}}^{(k)}$ are fully observed across all timesteps. This enables \emph{bidirectional conditioning}—realized through full self-attention, information flows forward from past and backward from future simultaneously:
\begin{equation}
    \boldsymbol{\epsilon}_\theta(\mathbf{Y}_{t,\ell}^{(k)}) = f_\theta\bigl(\mathbf{Y}_{t,1:L}^{(k)}, \mathbf{X}_{1:L}^{(k)}, \hat{\mathbf{Y}}_{1:L}^{(k)}, t\bigr), \quad \forall \ell \in \{1, \ldots, L\}
\end{equation}
This bidirectional flow facilitates the reconstruction of complex temporal dynamics that are difficult to infer from either direction alone.
For example, a peak can be identified when the forward view captures an upward trend before  a certain timestep and the backward view captures a downward trend immediately after it. 
The effect of the bidirectional flow is validated by the ablation results in Table~\ref{tab:main_results}.

%% file: 5-Experiment.tex
\section{Experiments}

\begin{table*}[t!]
\centering
\caption{Performance comparison across different datasets. Best results are in \textbf{bold}. $\uparrow$ indicates higher is better, $\downarrow$ indicates lower is better.}
\label{tab:main_results}
\renewcommand{\arraystretch}{1.2}
\resizebox{\textwidth}{!}{%
\begin{tabular}{l|ccc|ccc|ccc|ccc}
\toprule
\multirow{2}{*}{Model} & \multicolumn{3}{c|}{Streamflow} & \multicolumn{3}{c|}{Solar} & \multicolumn{3}{c|}{Temp} & \multicolumn{3}{c}{Methane} \\
\cmidrule{2-13}
 & NSE$\uparrow$ & RMSE$\downarrow$ & MAE$\downarrow$ & NSE$\uparrow$ & RMSE$\downarrow$ & MAE$\downarrow$ & NSE$\uparrow$ & RMSE$\downarrow$ & MAE$\downarrow$ & NSE$\uparrow$ & RMSE$\downarrow$ & MAE$\downarrow$ \\
\midrule
\multicolumn{13}{c}{\textit{Diffusion Baselines}} \\
\midrule

CSDI~\citep{tashiro2021csdi}             & $\dagger$ & 2.814 & 1.267 & 0.019 & 7.896 & 6.035 & $\dagger$ & 6.883 & 4.525 & $\dagger$ & 1.910 & 1.610 \\
\quad + $f_\omega$                       & 0.060 & 2.367 & 1.214 & 0.307 & 7.271 & 5.926 & 0.722 & 2.881 & 2.286 & $\dagger$ & 1.229 & 1.083 \\
SSSD~\citep{alcaraz2022diffusion}        & $\dagger$ & 2.583 & 1.161 & 0.446 & 5.930 & 4.468 & 0.605 & 4.002 & 3.303 & $\dagger$ & 0.974 & 0.890 \\
\quad + $f_\omega$                       & 0.071 & 2.348 & 1.201 & 0.374 & 6.534 & 4.865 & 0.746 & 2.756 & 2.113 & $\dagger$ & 1.083 & 0.986 \\
CSBI~\citep{chen2023csbi}               & $\dagger$ & 2.730 & 1.446 & 0.449 & 5.915 & 4.612 & 0.243 & 6.100 & 5.260 & $\dagger$ & 0.976 & 0.900 \\
\quad + $f_\omega$                       & 0.084 & 2.374 & 1.270 & 0.402 & 6.410 & 4.999 & 0.725 & 2.941 & 2.439 & $\dagger$ & 1.083 & 0.991 \\
DiffusionTS~\citep{yuan2024diffusionts}  & $\dagger$ & 3.230 & 1.670 & 0.026 & 7.852 & 5.980 & 0.293 & 5.671 & 4.461 & $\dagger$ & 1.661 & 1.399 \\
\quad + $f_\omega$                       & 0.033 & 2.446 & 1.269 & 0.259 & 7.229 & 5.965 & 0.747 & 2.920 & 2.213 & $\dagger$ & 1.222 & 1.068 \\
NsDiff~\citep{ye2025nsdiff}              & $\dagger$ & 3.907 & 2.686 & 0.515 & 5.546 & 4.556 & 0.700 & 2.902 & 2.371 & $\dagger$ & 1.597 & 1.372 \\
\quad + $f_\omega$                       & 0.105 & 2.488 & 1.325 & 0.473 & 6.131 & 4.987 & 0.763 & 2.562 & 2.063 & $\dagger$ & 1.262 & 1.085 \\
\midrule
\multicolumn{13}{c}{\textit{Ours}} \\
\midrule
$f_\omega$  ~\citep{kratzert2018rainfall}         & 0.131 & 2.139 & 1.099 & 0.345 & 6.407 & 5.280 & 0.813 & 2.306 & 1.848 & $\dagger$ & 1.074 & 0.940 \\
$\Phi_{\psi,\omega}$ & 0.481 & 1.679 & \textbf{0.725} & 0.486 & 5.707 & 4.565 & 0.852 & 2.205 & 1.752 & 0.481 & 0.371 & 0.280 \\
w/o prior            & $\dagger$ & 2.415 & 1.440 & 0.323 & 6.470 & 5.355 & 0.829 & 2.371 & 1.902 & $\dagger$ & 0.848 & 0.728 \\
w/o bidir.\ \& $w_k$ & 0.365 & 1.750 & 1.022 & 0.518 & 5.526 & 4.638 & 0.860 & 2.014 & 1.608 & 0.671 & 0.276 & 0.202 \\
w/o bidir.           & 0.463 & 1.576 & 0.903 & 0.519 & 5.519 & 4.636 & 0.859 & 2.010 & 1.600 & 0.623 & 0.276 & 0.203 \\
ZeroDiff             & \textbf{0.596} & \textbf{1.423} & 0.736 & \textbf{0.846} & \textbf{3.109} & \textbf{2.494} & \textbf{0.868} & \textbf{1.756} & \textbf{1.399} & \textbf{0.788} & \textbf{0.219} & \textbf{0.159} \\

\bottomrule
\multicolumn{13}{l}{\footnotesize $\dagger$ NSE $<$ 0 (cross-location transfer failed). $f_\omega$: LSTM~\citep{kratzert2018rainfall}; + $f_\omega$ = pre-trained on $f_\omega$ predictions.} \\
\multicolumn{13}{l}{\footnotesize bidir.\ = bidirectional denoiser (\S\,4.2); $w_k$ = moment-guided weighting (Eq.~\ref{eq:weights}).}
\end{tabular}%
}
\end{table*}

\begin{figure*}[t!]
    \centering
    \includegraphics[width=\textwidth]{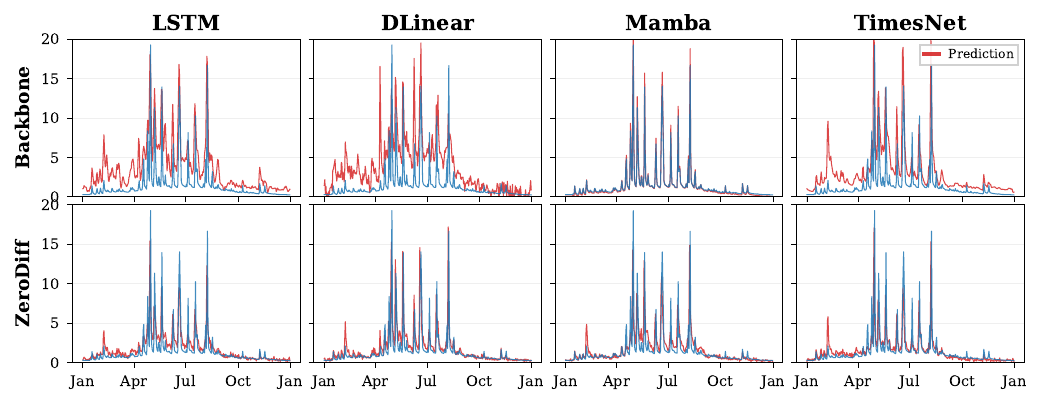}
    \caption{Qualitative comparison on Basin 02065500 (Year 1997, 365 days). Top row: reconstruction by each backbone dynamics model alone; bottom row: reconstruction enhanced by ZeroDiff. \textcolor{blue}{Blue}: ground truth streamflow; \textcolor{red}{red}: model prediction. ZeroDiff consistently improves temporal fidelity, particularly at peak flow events.}
    \label{fig:yearly_reconstruction}
\end{figure*}

\begin{table}[t!]
\centering
\caption{Effect of dynamics model architecture on CAMELS. Upper: dynamics model $f_\omega$ with moment estimation; lower: full ZeroDiff pipeline with each $f_\omega$.}
\label{tab:dynamic_models}
\renewcommand{\arraystretch}{1.2}
\begin{tabular}{l|ccc}
\toprule
Model & NSE$\uparrow$ & RMSE$\downarrow$ & MAE$\downarrow$ \\
\midrule
\multicolumn{4}{c}{\textit{Dynamics model $f_\omega$ only}} \\
\midrule
LSTM~\citep{kratzert2018rainfall}      & 0.481 & 1.679 & 0.725 \\
DLinear~\citep{zeng2023transformers}   & 0.428 & 1.777 & 0.818 \\
Mamba~\citep{gu2023mamba}    & 0.557 & 1.869 & 0.892 \\
TimesNet~\citep{wu2023timesnet}  & 0.517 & 1.613 & 0.706 \\
\midrule
\multicolumn{4}{c}{\textit{ZeroDiff with $f_\omega$}} \\
\midrule
ZeroDiff (LSTM)      & 0.596 & 1.423 & 0.736 \\
ZeroDiff (DLinear)   & 0.612 & 1.353 & 0.722 \\
ZeroDiff (Mamba)     & 0.667 & 1.618 & 0.828 \\
ZeroDiff (TimesNet)  & 0.593 & 1.427 & 0.724 \\
\bottomrule
\end{tabular}
\end{table}

\subsection{Experimental Setup}

\paragraph{Datasets.}

\begin{table}[H]
\centering
\caption{Dataset statistics. Each experiment masks one location and trains on the rest.}
\label{tab:dataset-statistics}
\renewcommand{\arraystretch}{1.2}
\resizebox{\columnwidth}{!}{%
\begin{tabular}{l|cccc}
\toprule
 & \textsc{Streamflow} & \textsc{Solar} & \textsc{Temp} & \textsc{Methane} \\
\midrule
Domain    & Hydrology & Energy & Hydrology & Carbon Emission \\
Exog. Dim & 33 & 48 & 8 & 16 \\
Length    & 4745 & 4015 & 1095 & 4015 \\
Test Loc. & 100 & 30 & 42 & 30 \\
\bottomrule
\end{tabular}%
}
\end{table}

We evaluate on four datasets spanning four scientific domains (Table~\ref{tab:dataset-statistics}). Unlike standard forecasting benchmarks that split data temporally, zero-shot reconstruction requires \emph{spatial} partitioning: we evaluate whether models trained on observed locations can generalize to held-out ones. For each dataset, we adopt a leave-one-location-out protocol—iteratively masking all target observations at one location while retaining its exogenous inputs, training on the remaining locations, and evaluating reconstruction at the held-out location. \textsc{Test Loc.} indicates how many independent trials are conducted per dataset; final metrics are averaged across all held-out locations.

CAMELS~\citep{newman2015development,addor2017camels} is a widely-used hydrology benchmark containing daily streamflow from 531 gauges across the contiguous US, with catchment attributes and meteorology for large-sample studies.
NHD contains daily stream water temperature derived from the National Hydrography Dataset~\citep{USGS2019NHD,USGS2024NationalHydrography} at approximately 1-km resolution.
Solar~\citep{mcgovern2015solar} is from the AMS 2013--2014 Solar Energy Prediction Contest, with daily solar energy from Oklahoma Mesonet stations and GEFS ensemble forecasts.
Methane~\citep{pastorello2020fluxnet2015,sun2025xmethanewet} provides daily methane (CH$_4$) flux from FLUXNET.

\paragraph{Fair Experiment}
Most baselines are designed for time series forecasting with look-back window $L_b$ and prediction horizon $H$. To fairly evaluate on reconstruction ($L=365$), we adopt two protocols. First, sliding window reconstruction: We partition the sequence into overlapping segments. For each segment, baselines can use ground-truth $\mathbf{Y}$ from observed locations as look-back and predict the subsequent horizon. Specifically, we use $L_b \in \{96, 192, 365\}$ and $H \in \{96, 192, 365\}$, sliding with stride $H$ to cover the full sequence. Final predictions are concatenated.
Second, exogenous-only input assumption: For methods supporting exogenous variables, we provide $\mathbf{Y}_{1:L}$ reconstructed from $\mathbf{X}_{1:L}$ as input, matching our zero-shot setting. For each baseline, we report the better of the two protocols, selected on a held-out validation set. Further details on experimental fairness are provided in Appendix~\ref{appendix:fair}.

\subsection{Main Experiments}
\paragraph{Baselines.}
Since no existing method directly addresses zero-shot time series reconstruction, we compare against representative diffusion-based approaches from two related tasks.
From \emph{time series imputation}: CSDI~\cite{tashiro2021csdi}, which performs conditional score-based diffusion with self-supervised masking; SSSD~\cite{alcaraz2022diffusion}, which combines structured state-space models with diffusion for long-range temporal modeling; and CSBI~\cite{chen2023csbi}, a Schr\"{o}dinger bridge-based method for conditional generation.
From \emph{time series forecasting}: Diffusion-TS~\cite{yuan2024diffusionts}, which uses an encoder-decoder Transformer with disentangled seasonal-trend representations; and NsDiff~\cite{ye2025nsdiff}, which introduces non-stationary diffusion to handle distribution shift.
All baselines require target observations during training; we adapt them using the sliding-window and exogenous-conditioned protocols described in Appendix~\ref{appendix:fair}, selecting the best configuration per baseline via validation.

\paragraph{Results.}
Table~\ref{tab:main_results} shows a clear pattern: standard diffusion models fail ($\dagger$) on Streamflow and Methane—datasets exhibiting high variability both across locations and within each time series. Without moment estimation to capture location-specific statistics, models trained on observed locations cannot generalize to unseen ones.

Providing baselines with $f_\omega$ predictions (rows + $f_\omega$) reduces errors. However, these models can only incorporate the prior through pre-training—using $f_\omega(\mathbf{X})$ as pseudo-labels. This treats the prior as training data rather than as an integral part of the diffusion process.

ZeroDiff instead integrates the informed prior directly into the forward process, reverse process, and optimization. The consistent gap between ZeroDiff and baselines (+ $f_\omega$) across all datasets demonstrates that tightly coupling the prior with diffusion is essential.

\paragraph{Ablation Study.}
We ablate ZeroDiff in the lower block of Table~\ref{tab:main_results}; effects are most pronounced on Streamflow where magnitudes differ by orders of magnitude across basins. Replacing $\mathcal{N}(\hat{\mathbf{Y}}, \sigma^2\mathbf{I})$ with $\mathcal{N}(\mathbf{0}, \mathbf{I})$ causes failure on Streamflow and Methane, while the deterministic prior $\Phi_{\psi,\omega}$ alone yields positive NSE everywhere; comparing $\Phi_{\psi,\omega}$ with ZeroDiff shows that diffusion calibration further improves performance by correcting systematic errors from dynamics learning. Removing the VAE collapses performance on high-variability datasets, as the shared dynamics model cannot rescale predictions without estimated $(\hat{\mu}, \hat{\sigma})$. Moment-guided weighting $w_k$ provides gains only when cross-location heterogeneity is high—on Solar and Temp where locations cluster in moment space, weighting becomes redundant. The bidirectional denoiser yields consistent improvement across all datasets: unlike forecasting where causality prohibits looking ahead, reconstruction allows information to flow both directions, enabling correction at sharp transitions.

\subsection{Choice of Dynamics Model}
The dynamics model $f_\omega$ is a plug-in component of ZeroDiff. In Table~\ref{tab:dynamic_models}, we evaluate four architectures on CAMELS (Streamflow), the most challenging dataset: LSTM~\citep{hochreiter1997long}, DLinear~\citep{zeng2023transformers}, Mamba~\citep{gu2023mamba}, and TimesNet~\citep{wu2023timesnet}.

As shown in Table~\ref{tab:dynamic_models}, ZeroDiff consistently improves all four backbones, confirming that the diffusion refinement is architecture-agnostic. Notably, although Mamba achieves the highest standalone NSE ($0.557$), our default choice is the simplest LSTM—yet ZeroDiff (LSTM) already reaches $0.596$, demonstrating that a correct formulation matters more than a powerful backbone. Even a naive dynamics model can capture sufficient temporal structure to provide a good prior for diffusion refinement. This also reveals the potential of ZeroDiff as a bridge between classical time series models and diffusion-based generation: practitioners can plug in any off-the-shelf backbone and obtain calibrated reconstructions. Figure~\ref{fig:yearly_reconstruction} illustrates this qualitatively—across all four backbones, ZeroDiff  tracks ground-truth peaks more faithfully than the backbone alone.

\subsection{Robustness to Prior Estimation Error}
\label{sec:prior_robustness}

The deterministic prior $\Phi_{\psi,\omega}$ varies substantially in quality across datasets and locations (Table~\ref{tab:main_results}), and the reverse posterior mean (Eq.~\ref{eq:posterior_mean}) anchors the diffusion trajectory to this prior through the $\gamma_2$ term. We analyze how the final reconstruction depends on prior quality from three angles: the role of the VAE, the behavior of diffusion under prior error, and the mechanism that decouples the two stages.

\begin{table}[t!]
\centering
\caption{Ablation of VAE moment estimation. NSE across four datasets.}
\label{tab:vae_necessity}
\renewcommand{\arraystretch}{1.2}
\resizebox{\columnwidth}{!}{%
\begin{tabular}{l|cccc}
\toprule
Pipeline & Solar & Temp & Streamflow & Methane \\
\midrule
VAE + LSTM + Diffusion        & 0.846 & 0.868 & 0.596 & 0.788 \\
LSTM + Diffusion (no VAE)     & 0.839 & 0.828 & 0.377 & $-0.655$ \\
\bottomrule
\end{tabular}%
}
\end{table}

\paragraph{Decoupling moment estimation.}
Zero-shot spatial extrapolation requires estimating both the marginal moments $(\mu, \sigma)$ and the temporal dynamics. The VAE decouples these two estimands, letting each component focus its generalization on one aspect. Table~\ref{tab:vae_necessity} shows that on low-heterogeneity datasets (Solar, Temp), removing the VAE causes negligible change, but on high-heterogeneity datasets it causes severe degradation, collapsing to $-0.655$ NSE on Methane. An imperfect prior is far better than no prior.

\begin{table}[t!]
\centering
\caption{Prior estimation error and diffusion recovery. \textsc{Prior NSE$<$0}: locations where the deterministic prior fails. \textsc{Improved}: locations where diffusion increases NSE.}
\label{tab:prior_recovery}
\renewcommand{\arraystretch}{1.2}
\resizebox{\columnwidth}{!}{%
\begin{tabular}{l|cccc}
\toprule
Dataset    & Max/Min ratio & VAE error & Prior NSE$<$0 & Improved \\
\midrule
Solar      & 1.3$\times$    & 2.0\%   & 0  & 32/32 \\
Temp       & 2.0$\times$    & 14.1\%  & 1  & 33/43 \\
Streamflow & 791$\times$    & 18.9\%  & 1  & 89/98 \\
Methane    & 728$\times$    & 9.9\%   & 4  & 28/31 \\
\bottomrule
\end{tabular}%
}
\end{table}

\paragraph{Diffusion recovery from prior error.}
The $\gamma_2$ anchoring term ties the reverse process to $\hat{\mathbf{Y}}$, raising the possibility that inaccurate moments propagate into the final reconstruction. Empirically, the opposite holds: the calibration gain is largest precisely where the prior is weakest. Comparing the $\Phi_{\psi,\omega}$ and ZeroDiff rows of Table~\ref{tab:main_results}, the absolute NSE improvement on Methane is $+0.307$ ($0.481 \to 0.788$), the largest among the four datasets. Table~\ref{tab:prior_recovery} confirms this at the location level: even on Methane, all 4 locations with negative prior NSE are recovered to positive NSE after diffusion, and 28 of 31 locations overall see improvement.

\begin{table}[t!]
\centering
\caption{Providing VAE-estimated moments $(\hat{\mu}, \hat{\sigma})$ as input features to the denoiser. Removing this conditioning consistently degrades performance.}
\label{tab:moment_conditioning}
\renewcommand{\arraystretch}{1.2}
\resizebox{\columnwidth}{!}{%
\begin{tabular}{l|cc}
\toprule
Dataset    & $(\hat{\mu}, \hat{\sigma})$ as input & $(\hat{\mu}, \hat{\sigma})$ removed \\
\midrule
Solar      & 0.846 & 0.831 \\
Temp       & 0.868 & 0.842 \\
Streamflow & 0.596 & 0.551 \\
Methane    & 0.788 & 0.731 \\
\bottomrule
\end{tabular}%
}
\end{table}

\paragraph{Moment-conditioned correction.}
The VAE-estimated moments $(\hat{\mu}, \hat{\sigma})$ are not only used to construct $\hat{\mathbf{Y}}$ via denormalization—they are also explicitly provided as input features to the denoiser: $\epsilon_\theta = f_\theta(\mathbf{Y}_t, [\mathbf{X}; \hat{\mu}; \hat{\sigma}], \hat{\mathbf{Y}}, t)$. During training, the denoiser observes a spectrum of $(\hat{\mu}, \hat{\sigma})$ values across observed locations, each paired with the corresponding ground-truth $\mathbf{Y}$, and learns a correction policy parameterized by these moments rather than a fixed offset relative to any specific prior. Table~\ref{tab:moment_conditioning} isolates this effect: removing $(\hat{\mu}, \hat{\sigma})$ from the denoiser's input consistently drops NSE, with the largest drops on high-heterogeneity datasets (Streamflow $-0.045$, Methane $-0.057$). Since the prior is constructed identically at observed and unobserved locations, this correction transfers to $\mathcal{U}$.

\subsection{Multi-Target Reconstruction}
\label{sec:multi_target}

In practical deployments such as continental-scale hydrological modeling, the set $\mathcal{U}$ of unobserved locations is large, and training a separate model per target is infeasible. ZeroDiff is designed to serve all targets with a single model: the moment-guided weighting (Eq.~\ref{eq:weights}) uses the centroid $(\hat{\mu}^{\ast}, \hat{\sigma}^{\ast})$ of estimated moments over $\mathcal{U}$ (Appendix~\ref{appendix:multi-target}), upweighting observed locations whose distributional characteristics are representative of the collective target set.

\begin{table}[t!]
\centering
\caption{Multi-Target reconstruction (NSE). Each row holds out a different number of locations per fold; a single model is trained to reconstruct all held-out targets simultaneously.}
\label{tab:multi_target}
\renewcommand{\arraystretch}{1.2}
\resizebox{\columnwidth}{!}{%
\begin{tabular}{c|cccc}
\toprule
Locations out & Streamflow & Solar & Temp & Methane \\
\midrule
1   & 0.596 & 0.846 & 0.868 & 0.788 \\
5   & 0.551 & 0.843 & 0.875 & 0.735 \\
10  & 0.554 & 0.843 & 0.865 & 0.749 \\
20  & 0.550 & 0.847 & 0.855 & 0.609 \\
\bottomrule
\end{tabular}%
}
\end{table}

Table~\ref{tab:multi_target} varies the number of held-out locations per fold from 1 to 20. On low-heterogeneity datasets (Solar, Temp), performance is essentially unchanged as $|\mathcal{U}|$ grows, indicating that the centroid is an effective summary when targets cluster in moment space. On high-heterogeneity datasets, NSE drops modestly (Streamflow $0.596 \to 0.550$, Methane $0.788 \to 0.609$) as the centroid must compromise across more diverse targets. This degradation is a deliberate trade-off: one trained model serves all $|\mathcal{U}|$ locations, providing an $|\mathcal{U}|$-fold reduction in computational cost relative to per-location training, with the per-location accuracy gap remaining small when targets share a geographic region.

%% file: 6-Conclusion.tex
\section{Conclusion}

ZeroDiff enables zero-shot time series reconstruction by moment estimation and dynamics learning, constructing an informed prior from exogenous variables alone, then calibrating systematic errors through diffusion. The calibration learned on observed locations generalizes to unobserved ones. Experiments across four real-world datasets show that ZeroDiff significantly outperforms existing diffusion methods, particularly on datasets with high cross-location variability where baselines fail entirely.

%% file: 7-Appendix.tex
\newpage
\section{Algorithms for Training and Inference}
\label{sec:training}

\begin{algorithm}[H]
   \caption{Training}
   \label{alg:training}
\begin{algorithmic}
   \STATE {\bfseries Input:} $\mathcal{O}$, $\mathcal{U}$, $\{\mathbf{X}^{(k)}\}$, $\{\mathbf{Y}^{(k)}\}_{k \in \mathcal{O}}$, diffusion steps $T$
   \STATE Train VAE on $\mathcal{O}$: encoder $q_\psi$, feature encoder $\phi$, decoder $\mathrm{Dec}_\psi$ \hfill \textit{// cross-modal moment estimation}
   \STATE For all $k$: $(\hat{\mu}_k, \hat{\sigma}_k) \gets \mathrm{Dec}_\psi(\mathbf{0}, \phi(\mathbf{Z}^{(k)}))$ \hfill \textit{// infer magnitude from $\mathbf{X}$}
   \STATE Train dynamics model $f_\omega$ on $\mathcal{O}$ in standardized space \hfill \textit{// shared response pattern}
   \STATE Compute $(\hat{\mu}^*, \hat{\sigma}^*) \gets \frac{1}{|\mathcal{U}|} \sum_{j \in \mathcal{U}} (\hat{\mu}_j, \hat{\sigma}_j)$ \hfill \textit{// target centroid in moment space}
   \REPEAT
   \STATE Sample $k \in \mathcal{O}$ with probability $\propto w_k$ \hfill \textit{// moment-guided weighting}
   \STATE Compute $\hat{\mathbf{Y}}^{(k)} \gets \hat{\mu}_k + \hat{\sigma}_k \cdot f_\omega(\mathbf{X}^{(k)})$ \hfill \textit{// pattern + magnitude}
   \STATE Sample $t \sim \mathrm{Uniform}(\{1, \dots, T\})$, $\boldsymbol{\epsilon} \sim \mathcal{N}(\mathbf{0}, \mathbf{I})$
   \STATE $\mathbf{Y}_t^{(k)} \gets \sqrt{\bar{\alpha}_t} \mathbf{Y}^{(k)} + (1 - \sqrt{\bar{\alpha}_t}) \hat{\mathbf{Y}}^{(k)} + \sqrt{\bar{\sigma}_t} \boldsymbol{\epsilon}$ \hfill \textit{// diffuse toward prior}
   \STATE Update $\theta$ via $\nabla_\theta w_k \|\boldsymbol{\epsilon} - \boldsymbol{\epsilon}_\theta(\mathbf{Y}_t^{(k)}, \mathbf{X}^{(k)}, \hat{\mathbf{Y}}^{(k)}, t)\|^2$ \hfill \textit{// calibration on $\mathcal{O}$}
   \UNTIL{convergence}
\end{algorithmic}
\end{algorithm}

\begin{algorithm}[H]
   \caption{Inference}
   \label{alg:inference}
\begin{algorithmic}
   \STATE {\bfseries Input:} $j \in \mathcal{U}$, $\mathbf{X}^{(j)}$, trained VAE ($\phi$, $\mathrm{Dec}_\psi$), dynamics model $f_\omega$, denoiser $\boldsymbol{\epsilon}_\theta$
   \STATE $(\hat{\mu}_j, \hat{\sigma}_j) \gets \mathrm{Dec}_\psi(\mathbf{0}, \phi(\mathbf{Z}^{(j)}))$ \hfill \textit{// zero-shot moment estimation}
   \STATE $\hat{\mathbf{Y}}^{(j)} \gets \hat{\mu}_j + \hat{\sigma}_j \cdot f_\omega(\mathbf{X}^{(j)})$ \hfill \textit{// pattern + magnitude}
   \STATE $\mathbf{Y}_T^{(j)} \sim \mathcal{N}(\hat{\mathbf{Y}}^{(j)}, \bar\sigma_T \mathbf{I})$ \hfill \textit{// warm-start from prior, not noise}
   \FOR{$t = T$ {\bfseries to} $1$}
   \STATE $\boldsymbol{\epsilon}_\theta \gets f_\theta(\mathbf{Y}_t^{(j)}, \mathbf{X}^{(j)}, \hat{\mathbf{Y}}^{(j)}, t)$ \hfill \textit{// bidirectional conditioning}
   \STATE $\hat{\mathbf{Y}}_0^{(j)} \gets \frac{1}{\sqrt{\bar{\alpha}_t}}\bigl(\mathbf{Y}_t^{(j)} - (1 - \sqrt{\bar{\alpha}_t})\hat{\mathbf{Y}}^{(j)} - \sqrt{\bar{\sigma}_t}\boldsymbol{\epsilon}_\theta\bigr)$ \hfill \textit{// reparameterization}
   \STATE $\boldsymbol{\mu}_{t-1}^{(j)} \gets \gamma_0 \hat{\mathbf{Y}}_0^{(j)} + \gamma_1 \mathbf{Y}_t^{(j)} + \gamma_2 \hat{\mathbf{Y}}^{(j)}$ \hfill \textit{// three-term posterior}
   \STATE $\mathbf{Y}_{t-1}^{(j)} \gets \boldsymbol{\mu}_{t-1}^{(j)} + \sigma_t \mathbf{z}$, \quad $\mathbf{z} \sim \mathcal{N}(\mathbf{0}, \mathbf{I})$ if $t > 1$
   \ENDFOR
   \STATE {\bfseries Output:} $\mathbf{Y}_0^{(j)}$
\end{algorithmic}
\end{algorithm}

\noindent\textbf{Remark.} Unlike standard DDPM that initializes from
$\mathcal{N}(\mathbf{0}, \mathbf{I})$, our method performs warm-start
diffusion from the informed prior $\hat{\mathbf{Y}}^{(j)}$. The prior
anchors the reverse process at three levels: (i) initialization of
$\mathbf{Y}_T^{(j)}$, (ii) the reparameterization of
$\hat{\mathbf{Y}}_0^{(j)}$, and (iii) the $\gamma_2$ term in the posterior
mean. This transforms diffusion from generation to
calibration---refining a reasonable estimate rather than constructing
from noise. The coefficients $\gamma_0, \gamma_1, \gamma_2$, the variance
schedule $\bar\sigma_t$, and the three-term posterior mean are derived in
Appendix~\ref{appendix:posterior_derivation}.

\newpage
\section{Bayesian Derivation of the Posterior Distribution}
\label{appendix:posterior_derivation}

We derive the posterior distribution for our informed-prior diffusion process. For notational simplicity, we omit the location index and write $\mathbf{Y}$ for the target, $\hat{\mathbf{Y}}$ for the informed prior, and $\mathbf{X}$ for the exogenous input. Unlike standard DDPM which diffuses toward pure noise $\mathbf{Y}_T \sim \mathcal{N}(\mathbf{0}, \mathbf{I})$, we diffuse toward the informed prior $\hat{\mathbf{Y}}$, enabling warm-start initialization for calibration.

\subsection{Forward Process}

The one-step forward transition is:
\begin{equation}
\label{eq:app_forward}
    \mathbf{Y}_t = \sqrt{\alpha_t} \mathbf{Y}_{t-1} + (1 - \sqrt{\alpha_t}) \hat{\mathbf{Y}} + \sqrt{\beta_t} \boldsymbol{\epsilon}_t, \quad \boldsymbol{\epsilon}_t \sim \mathcal{N}(\mathbf{0}, \mathbf{I})
\end{equation}
where $\alpha_t := 1 - \beta_t$, $\bar\alpha_t := \prod_{i=1}^t \alpha_i$, and $\bar\sigma_t := 1 - \bar\alpha_t$. Unrolling recursively:
\begin{align}
\mathbf{Y}_t &= \sqrt{\alpha_t} \mathbf{Y}_{t-1} + (1 - \sqrt{\alpha_t}) \hat{\mathbf{Y}} + \sqrt{\beta_t} \boldsymbol{\epsilon}_t \notag \\
&= \sqrt{\alpha_t} \left[ \sqrt{\alpha_{t-1}} \mathbf{Y}_{t-2} + (1 - \sqrt{\alpha_{t-1}}) \hat{\mathbf{Y}} + \sqrt{\beta_{t-1}} \boldsymbol{\epsilon}_{t-1} \right] + (1 - \sqrt{\alpha_t}) \hat{\mathbf{Y}} + \sqrt{\beta_t} \boldsymbol{\epsilon}_t \notag \\
&= \sqrt{\alpha_t \alpha_{t-1}} \mathbf{Y}_{t-2} + \left(1 - \sqrt{\alpha_t \alpha_{t-1}}\right) \hat{\mathbf{Y}} + \sqrt{\alpha_t \beta_{t-1} + \beta_t} \boldsymbol{\epsilon} \notag \\
&\;\vdots \notag \\
&= \sqrt{\bar\alpha_t} \mathbf{Y} + (1 - \sqrt{\bar\alpha_t}) \hat{\mathbf{Y}} + \sqrt{\bar\sigma_t} \boldsymbol{\epsilon} \label{eq:app_cumulative}
\end{align}
Thus $q(\mathbf{Y}_t | \mathbf{Y}, \hat{\mathbf{Y}}) = \mathcal{N}\left(\sqrt{\bar\alpha_t} \mathbf{Y} + (1 - \sqrt{\bar\alpha_t}) \hat{\mathbf{Y}}, \bar\sigma_t \mathbf{I}\right)$, and at $t=T$: $\mathbf{Y}_T \sim \mathcal{N}(\hat{\mathbf{Y}}, \bar\sigma_T \mathbf{I})$—the warm-start initialization centered at the prior.

\subsection{Reverse Posterior Derivation}

We derive $q(\mathbf{Y}_{t-1} | \mathbf{Y}_t, \mathbf{Y}, \hat{\mathbf{Y}})$ via Bayes' rule:
\begin{equation}
q(\mathbf{Y}_{t-1} | \mathbf{Y}_t, \mathbf{Y}, \hat{\mathbf{Y}}) \propto q(\mathbf{Y}_t | \mathbf{Y}_{t-1}, \hat{\mathbf{Y}}) \cdot q(\mathbf{Y}_{t-1} | \mathbf{Y}, \hat{\mathbf{Y}})
\end{equation}

From Eq.~\eqref{eq:app_forward}, the transition density is:
\begin{equation}
q(\mathbf{Y}_t | \mathbf{Y}_{t-1}, \hat{\mathbf{Y}}) = \mathcal{N}\left(\sqrt{\alpha_t} \mathbf{Y}_{t-1} + (1 - \sqrt{\alpha_t}) \hat{\mathbf{Y}}, \beta_t \mathbf{I}\right)
\end{equation}

From Eq.~\eqref{eq:app_cumulative} at $t-1$:
\begin{equation}
q(\mathbf{Y}_{t-1} | \mathbf{Y}, \hat{\mathbf{Y}}) = \mathcal{N}\left(\sqrt{\bar\alpha_{t-1}} \mathbf{Y} + (1 - \sqrt{\bar\alpha_{t-1}}) \hat{\mathbf{Y}}, \bar\sigma_{t-1} \mathbf{I}\right)
\end{equation}

Define $\mathbf{A} := \mathbf{Y}_t - (1 - \sqrt{\alpha_t}) \hat{\mathbf{Y}}$ and $\mathbf{B} := \sqrt{\bar\alpha_{t-1}} \mathbf{Y} + (1 - \sqrt{\bar\alpha_{t-1}}) \hat{\mathbf{Y}}$. The product of two Gaussians gives:
\begin{align}
&q(\mathbf{Y}_{t-1} | \mathbf{Y}_t, \mathbf{Y}, \hat{\mathbf{Y}}) \propto \exp\left( -\frac{(\mathbf{A} - \sqrt{\alpha_t} \mathbf{Y}_{t-1})^2}{2\beta_t} - \frac{(\mathbf{Y}_{t-1} - \mathbf{B})^2}{2\bar\sigma_{t-1}} \right) \notag \\
&= \exp\left( -\frac{1}{2} \left[ \left(\frac{\alpha_t}{\beta_t} + \frac{1}{\bar\sigma_{t-1}}\right) \mathbf{Y}_{t-1}^2 - 2\left(\frac{\sqrt{\alpha_t} \mathbf{A}}{\beta_t} + \frac{\mathbf{B}}{\bar\sigma_{t-1}}\right) \mathbf{Y}_{t-1} + \text{const} \right] \right)
\end{align}

Completing the square in $\mathbf{Y}_{t-1}$, the posterior is Gaussian with:
\begin{align}
\tilde{\sigma}_{t-1} &= \left(\frac{\alpha_t}{\beta_t} + \frac{1}{\bar\sigma_{t-1}}\right)^{-1} = \frac{\beta_t \bar\sigma_{t-1}}{\alpha_t \bar\sigma_{t-1} + \beta_t} \label{eq:app_posterior_var} \\
\tilde{\mu}_{t-1} &= \tilde{\sigma}_{t-1} \left(\frac{\sqrt{\alpha_t} \mathbf{A}}{\beta_t} + \frac{\mathbf{B}}{\bar\sigma_{t-1}}\right) = \frac{\sqrt{\alpha_t} \bar\sigma_{t-1} \mathbf{A} + \beta_t \mathbf{B}}{\alpha_t \bar\sigma_{t-1} + \beta_t} \label{eq:app_posterior_mean_raw}
\end{align}

Substituting $\mathbf{A} = \mathbf{Y}_t - (1 - \sqrt{\alpha_t}) \hat{\mathbf{Y}}$ and $\mathbf{B} = \sqrt{\bar\alpha_{t-1}} \mathbf{Y} + (1 - \sqrt{\bar\alpha_{t-1}}) \hat{\mathbf{Y}}$ into Eq.~\eqref{eq:app_posterior_mean_raw}:
\begin{align}
\tilde{\mu}_{t-1} &= \frac{\sqrt{\alpha_t} \bar\sigma_{t-1} \left[\mathbf{Y}_t - (1 - \sqrt{\alpha_t}) \hat{\mathbf{Y}}\right] + \beta_t \left[\sqrt{\bar\alpha_{t-1}} \mathbf{Y} + (1 - \sqrt{\bar\alpha_{t-1}}) \hat{\mathbf{Y}}\right]}{\alpha_t \bar\sigma_{t-1} + \beta_t} \notag \\
&= \frac{\sqrt{\bar\alpha_{t-1}} \beta_t}{\alpha_t \bar\sigma_{t-1} + \beta_t} \mathbf{Y} + \frac{\sqrt{\alpha_t} \bar\sigma_{t-1}}{\alpha_t \bar\sigma_{t-1} + \beta_t} \mathbf{Y}_t + \frac{\sqrt{\alpha_t}(\sqrt{\alpha_t} - 1) \bar\sigma_{t-1} + (1 - \sqrt{\bar\alpha_{t-1}}) \beta_t}{\alpha_t \bar\sigma_{t-1} + \beta_t} \hat{\mathbf{Y}}
\end{align}

Thus the posterior mean has the \textbf{three-term form}:
\begin{equation}
\label{eq:app_posterior_mean}
\boxed{\tilde{\mu}_{t-1} = \gamma_0 \mathbf{Y} + \gamma_1 \mathbf{Y}_t + \gamma_2 \hat{\mathbf{Y}}}
\end{equation}
where:
\begin{align}
\gamma_0 &= \frac{\sqrt{\bar\alpha_{t-1}} \beta_t}{\alpha_t \bar\sigma_{t-1} + \beta_t}, \quad
\gamma_1 = \frac{\sqrt{\alpha_t} \bar\sigma_{t-1}}{\alpha_t \bar\sigma_{t-1} + \beta_t}, \quad
\gamma_2 = \frac{\sqrt{\alpha_t}(\sqrt{\alpha_t} - 1) \bar\sigma_{t-1} + (1 - \sqrt{\bar\alpha_{t-1}}) \beta_t}{\alpha_t \bar\sigma_{t-1} + \beta_t} \label{eq:app_gammas}
\end{align}

This differs from standard DDPM where $\tilde{\mu}_{t-1}^{\text{DDPM}} = \gamma_0 \mathbf{Y} + \gamma_1 \mathbf{Y}_t$. The additional $\gamma_2 \hat{\mathbf{Y}}$ term anchors each denoising step to the prior prediction.

\subsection{Inference via Reparameterization}

During inference, $\mathbf{Y}$ is unknown. From Eq.~\eqref{eq:app_cumulative}, we estimate:
\begin{equation}
\hat{\mathbf{Y}}_0 = \frac{1}{\sqrt{\bar\alpha_t}} \left( \mathbf{Y}_t - (1 - \sqrt{\bar\alpha_t}) \hat{\mathbf{Y}} - \sqrt{\bar\sigma_t} \boldsymbol{\epsilon}_\theta(\mathbf{Y}_t, \mathbf{X}, \hat{\mathbf{Y}}, t) \right)
\end{equation}
where $\boldsymbol{\epsilon}_\theta$ is the learned denoiser. The reverse step is:
\begin{equation}
\mathbf{Y}_{t-1} = \gamma_0 \hat{\mathbf{Y}}_0 + \gamma_1 \mathbf{Y}_t + \gamma_2 \hat{\mathbf{Y}} + \sqrt{\tilde{\sigma}_{t-1}} \mathbf{z}, \quad \mathbf{z} \sim \mathcal{N}(\mathbf{0}, \mathbf{I})
\end{equation}

The prior $\hat{\mathbf{Y}}$ anchors the process at three levels: (1) initialization $\mathbf{Y}_T \sim \mathcal{N}(\hat{\mathbf{Y}}, \bar\sigma_T \mathbf{I})$, (2) reparameterization of $\hat{\mathbf{Y}}_0$, and (3) the $\gamma_2 \hat{\mathbf{Y}}$ term in the posterior mean. This triple anchoring transforms diffusion from generation to calibration: the network learns residual corrections rather than the full data distribution.

\clearpage
\section{Case Study: Full-Series Reconstruction on an Ungauged Basin}
\begin{figure}[H]
    \centering
    \includegraphics[width=\textwidth]{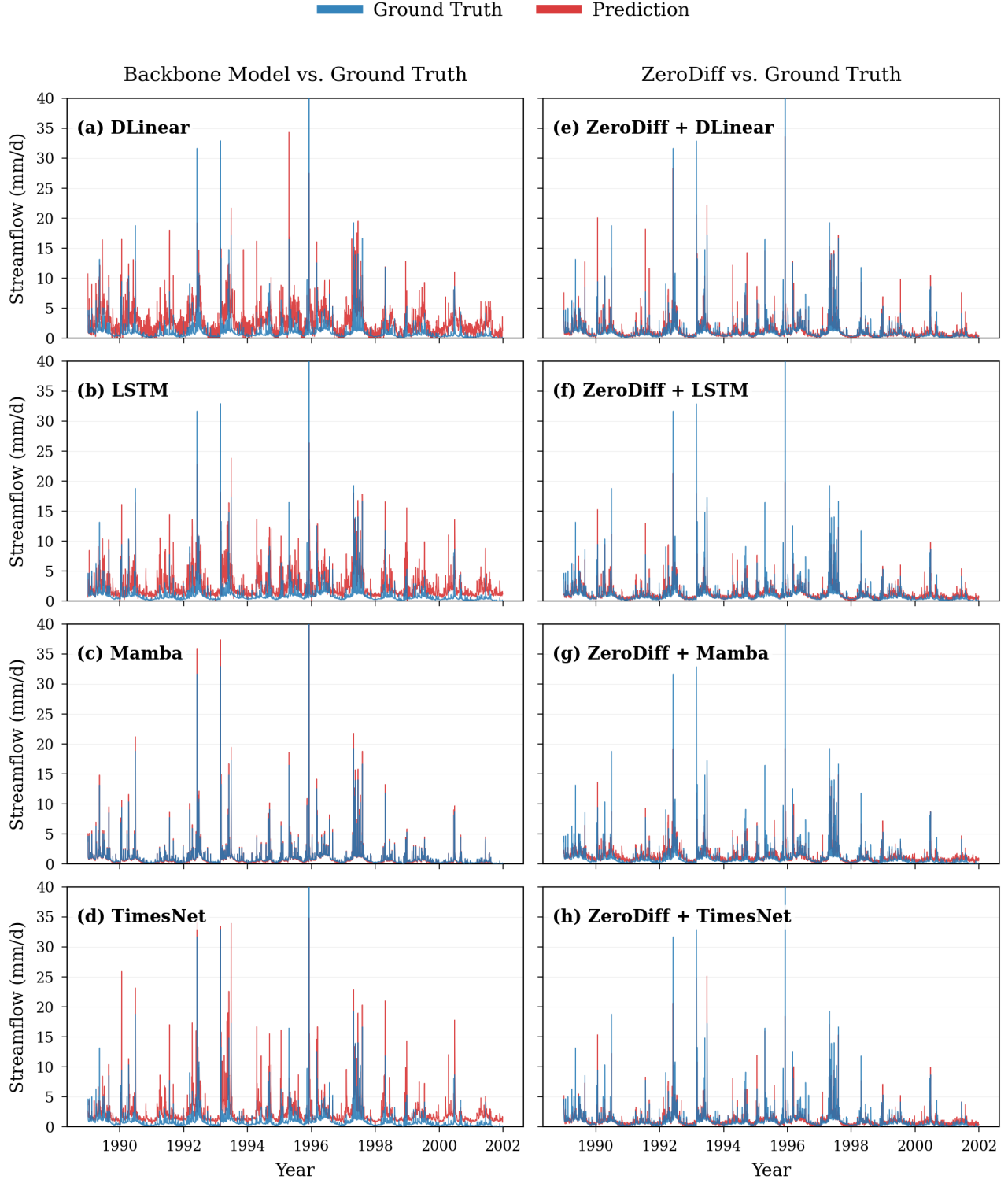}
    \caption{
Full time series reconstruction for Basin 02065500 over the entire 13-year
training period (1989--2001, 4{,}745 days).
    }
    \label{fig:full_timeseries_reconstruction}
\end{figure}

\newpage
\section{Dataset Details}
\label{appendix:dataset}

We describe the four datasets, their exogenous variables, and train/test splits.
\subsection{Dataset Overview}

\begin{table}[h]
\centering
\caption{Dataset statistics: temporal coverage, train/test splits, and exogenous variable counts. All datasets use daily resolution.}
\label{tab:dataset-full}
\begin{tabular}{l c c c c c c c}
\toprule
\textsc{Dataset} & \textsc{Full Period} & \textsc{Train Period} & \textsc{Days} & \textsc{Test Loc.} & \textsc{Dynamic} & \textsc{Static} & \textsc{Exog.} \\
\midrule
CAMELS    & 1989--2009 & 1989--2001 & 4,745 & 100 & 6  & 27 & 33 \\
Temp  & 2010--2012 & 2010--2012 & 1,095 & 42  & 8  & 0  & 8 \\
Solar     & 1994--2007 & 1994--2004 & 4,015 & 30  & 45 & 3  & 48 \\
Methane      & 2008--2018 & 2008--2018 & 4,015 & 30  & 7  & 9  & 16 \\
\bottomrule
\end{tabular}
\end{table}

\subsection{Dataset Partition}

The data splitting strategy for zero-shot reconstruction differs fundamentally from time series forecasting. While forecasting typically partitions data along the temporal dimension, our task requires spatial partitioning: for each experiment, we mask all target observations at a held-out location while retaining its exogenous variables, then leverage data from remaining locations to reconstruct the masked targets.

We reserve a subset of locations as the validation set. Although their targets are masked during training, we use their reconstruction performance to tune hyperparameters. The remaining held-out locations form the test set for final evaluation. This spatial cross-validation ensures that hyperparameters are selected without leaking information from test locations.

For most baselines, we do not adopt the default hyperparameters reported in their original papers, as these were optimized for forecasting tasks. Instead, we conduct task-specific hyperparameter search on our validation set to ensure fair comparison. 

\paragraph{Leave-One-Location-Out Evaluation.}
We adopt a leave-one-location-out protocol to evaluate zero-shot generalization. For each dataset, we iteratively hold out one target location: all target observations at this location are masked while exogenous variables remain available. The model is trained on the remaining locations and evaluated on its ability to reconstruct the held-out target series. This procedure is repeated for all target locations—100 times for CAMELS, 42 for NHD, and 30 for both Solar and Methane. Final metrics are averaged across all held-out locations to provide a robust estimate of zero-shot reconstruction performance.

\clearpage
\section{Efficiency}

\subsection{Multi-Target Reconstruction}
\label{appendix:multi-target}
 
A natural question arises: why not train a separate model for each target location $j \in \mathcal{U}$? Location-specific training would allow moment-guided weighting to focus entirely on $(\hat{\mu}_j, \hat{\sigma}_j)$, potentially yielding better reconstruction for that specific location.
 
While this approach maximizes per-location performance, it becomes impractical when $|\mathcal{U}|$ is large. Training and storing separate models for each target location incurs computational cost linear in $|\mathcal{U}|$—prohibitive in applications such as continental-scale hydrological modeling where thousands of ungauged locations require reconstruction.
 
Our framework addresses this via the moment centroid. Let $(\hat{\mu}^*, \hat{\sigma}^*)$ denote the centroid of estimated statistics over $j \in \mathcal{U}$:
\begin{equation}
\label{eq:moment_centroid}
    (\hat{\mu}^*, \hat{\sigma}^*) = \frac{1}{|\mathcal{U}|} \sum_{j \in \mathcal{U}} (\hat{\mu}_j, \hat{\sigma}_j)
\end{equation}
By defining $(\hat{\mu}^*, \hat{\sigma}^*)$ as the geometric center over all $j \in \mathcal{U}$, a single model serves multiple target locations simultaneously. The moment-guided weighting $w_k$ (Eq.~\ref{eq:weights}) then upweights training locations whose distributions are representative of the \emph{collective} target set, rather than any individual target.
 
This design reflects a deliberate trade-off: we accept a modest reduction in per-location accuracy in exchange for an $|\mathcal{U}|$-fold improvement in efficiency. In practice, when target locations cluster in moment space—as is common within a geographic region—the centroid closely approximates individual targets, and the performance gap is minimal. Table~\ref{tab:multi_target} in the main text validates this empirically: as $|\mathcal{U}|$ grows from 1 to 20, performance remains nearly unchanged on low-heterogeneity datasets (Solar, Temp) and degrades only modestly on high-heterogeneity datasets, while training cost is reduced by a factor of $|\mathcal{U}|$.
 
For applications requiring maximum accuracy at a specific location, one can set $\mathcal{U} = \{j\}$ and train a location-specific model. Our framework supports both modes: efficient batch reconstruction when throughput matters, and focused single-target reconstruction when accuracy is paramount.

\clearpage
\subsection{Diffusion Computational Efficiency Analysis}
\begin{table*}[h]
\centering
\caption{Computational efficiency of six diffusion-based imputation models across all datasets.
Models are sorted by total per-fold time (ascending) within each panel.
All datasets use 365-day sequences. Training uses early stopping with patience of 20 epochs.
All experiments conducted on H100.}
\label{tab:efficiency_all}
\vspace{4pt}
\renewcommand{\arraystretch}{1.15}
\begin{tabular}{l r c c c c c c}
\toprule
\textbf{Model}
  & \textbf{\#Params}
  & $T$
  & $N_s$
  & \textbf{Train (min)}
  & \textbf{Infer (min)}
  & \textbf{ms/sample}
  & \textbf{Total (min)} \\
\midrule
\multicolumn{8}{l}{\textbf{(a) Streamflow} \;\textit{(531 locations, 33 features, 6{,}903 samples/split)}} \\[2pt]
ZeroDiff              & 271\,K  & 20  & 100 & \textbf{4.3}  & \underline{9.1}   & \underline{39.4}  & \textbf{13.4}    \\
DiffusionTS$^\dagger$ & 2.33\,M & 100 & 1   & 12.1          & \textbf{3.0}      & \textbf{21.4}     & \underline{15.1} \\
NsDiff                & 270\,K  & 20  & 100 & 16.8          & 9.2               & 39.9              & 26.0             \\
CSDI                  & 618\,K  & 50  & 10  & 23.0          & 44.4              & 192.9             & 67.4             \\
SSSD$^\ddagger$       & 29.4\,M & 200 & 10  & 12.2          & 165.8             & 720.5             & 178.0            \\
CSBI                  & 137\,K  & 100 & 20  & 19.1          & 160.0             & 695.4             & 179.1            \\
\addlinespace[6pt]
\multicolumn{8}{l}{\textbf{(b) Solar} \;\textit{(500 locations, 48 features, 5{,}500 samples/split)}} \\[2pt]
DiffusionTS$^\dagger$ & 2.32\,M & 100 & 1   & \underline{6.3} & \textbf{4.0}    & \textbf{22.0}     & \textbf{10.3}    \\
ZeroDiff              & 270\,K  & 20  & 100 & \textbf{4.2}  & \underline{7.2}   & \underline{39.5}  & \underline{11.4} \\
NsDiff                & 269\,K  & 20  & 100 & 12.4          & 7.3               & 39.9              & 19.7             \\
CSDI                  & 617\,K  & 50  & 10  & 12.3          & 35.4              & 192.9             & 47.7             \\
CSBI                  & 137\,K  & 100 & 20  & 15.2          & 127.2             & 693.8             & 142.4            \\
SSSD$^\ddagger$       & 29.4\,M & 200 & 10  & 10.8          & 132.4             & 722.2             & 143.2            \\
\addlinespace[6pt]
\multicolumn{8}{l}{\textbf{(c) Temp } \;\textit{(45 locations, 8 features, 135 samples/split)}} \\[2pt]
ZeroDiff              & 269\,K  & 20  & 100 & \textbf{0.6}  & \underline{0.5}   & 77.0              & \textbf{1.1}     \\
NsDiff                & 269\,K  & 20  & 100 & \underline{0.8} & 0.5              & \underline{76.9}  & \underline{1.3}  \\
DiffusionTS$^\dagger$ & 2.32\,M & 100 & 1   & 1.8           & \textbf{0.3}      & \textbf{41.0}     & 2.1              \\
CSDI                  & 617\,K  & 50  & 10  & 1.6           & 2.2               & 332.1             & 3.8              \\
SSSD$^\ddagger$       & 29.4\,M & 200 & 10  & 1.0           & 5.2               & 770.0             & 6.2              \\
CSBI                  & 137\,K  & 100 & 20  & 1.0           & 12.3              & 1{,}821.0         & 13.3             \\
\bottomrule
\end{tabular}
\vspace{4pt}
\end{table*}

\newpage
\section{Fairness Details}
\label{appendix:fair}
Most baselines are designed for time series forecasting with look-back window $L_b$ and prediction horizon $H$. To fairly evaluate on reconstruction ($L=365$), we adopt two protocols.

\subsection{Sliding Window Reconstruction} We partition the sequence into segments where baselines use ground-truth $\mathbf{Y}$ from observed locations as look-back to predict subsequent horizons. We consider $L_b, H \in \{96, 192, 365\}$, including the $L_b=H=365$ setting where the model reconstructs the current window conditioned on itself as a calibration baseline. Predictions are concatenated with stride $H$ to cover the full sequence. The optimal $(L_b, H)$ for each baseline is selected via validation set performance and reported in Table~\ref{tab:main_results}.

\subsection{Exogenous-Conditioned Reconstruction} For methods supporting exogenous inputs, we provide the informed prior $\hat{\mathbf{Y}}_{1:L}$ reconstructed from $\mathbf{X}_{1:L}$ as the look-back window, matching our zero-shot setting where no ground-truth targets are available. This protocol evaluates whether baselines can refine an exogenous-based estimate.

Note that protocol (i) provides baselines with oracle access to target observations—an advantage unavailable in true zero-shot scenarios.

\subsection{Hyperparameter Tuning}
For zero-shot time series reconstruction, dataset splitting differs fundamentally from time series forecasting. While forecasting typically partitions data along the temporal dimension, our task partitions along the spatial dimension. In each experiment, we mask all target observations at a held-out location while retaining its exogenous variables, then leverage data from other locations to perform zero-shot prediction of the masked target variable. We reserve several locations as a validation set; although their targets are masked during inference, we use their ground-truth values for hyperparameter selection.

For most baselines, we do not use the default hyperparameters reported in their original papers, as those were tuned for time series forecasting rather than our reconstruction task. Instead, we conduct task-specific hyperparameter search. Our experiments reveal that among $L_b, H \in \{96, 192, 365\}$, the configuration $L_b=H=365$ performs best for most baselines—reconstructing a full year from a full year of context. We attribute this to the misalignment artifacts that arise when $L_b$ or $H$ take smaller values, as discontinuous segments must be concatenated to cover the annual sequence.

\clearpage
\section{Gradient Isolation vs.\ Joint Optimization of the Prior and Denoiser}
\label{appendix:end2end}

ZeroDiff trains the prior construction $\Phi_{\psi,\omega}$ and the
diffusion denoiser $\boldsymbol{\epsilon}_\theta$ in two stages
(Algorithm~\ref{alg:training}), so that no gradient flows between the two
components. A natural alternative is joint (end-to-end) optimization,
where the diffusion loss backpropagates into the dynamics model through a
differentiable denormalization--renormalization chain. We evaluate this
variant on all four datasets, initializing from the two-stage solution
and jointly fine-tuning with a learning rate of $10^{-5}$ for the prior
components and $10^{-3}$ for the denoiser.

\begin{table}[H]
\centering
\caption{Two-stage training vs.\ joint optimization (NSE) across all four
datasets.}
\label{tab:end2end}
\renewcommand{\arraystretch}{1.2}
\begin{tabular}{l|ccc}
\toprule
Dataset    & Two-stage & Joint & $\Delta$ \\
\midrule
Solar      & 0.846 & 0.855 & $+0.009$ \\
Temp       & 0.868 & 0.886 & $+0.018$ \\
Streamflow & 0.596 & 0.631 & $+0.035$ \\
Methane    & 0.788 & 0.758 & $-0.030$ \\
\bottomrule
\end{tabular}
\end{table}

Table~\ref{tab:end2end} reveals a dataset-dependent trade-off. Under
gradient isolation, the prior construction solves a pure
$\mathbf{X}\to\mathbf{Y}$ regression problem, independent of the denoiser
$\boldsymbol{\epsilon}_\theta$. Under joint optimization, its role
changes: it is optimized under the diffusion loss and can drift toward
producing priors that a \emph{specific} $\mathcal{O}$-trained denoiser
finds easy to calibrate. The two components can thus co-adapt on observed
locations $\mathcal{O}$ in ways that are not part of the underlying
$\mathbf{X}\to\mathbf{Y}$ relationship, and such co-adaptation does not
necessarily transfer to unobserved locations $\mathcal{U}$.

How much room exists for co-adaptation depends on how strongly the data
constrain the prior. When the $\mathbf{X}\to\mathbf{Y}$ coupling is strong
(Solar, Temp, Streamflow), the regression signal pulls the prior toward a
narrow band of solutions, and joint optimization delivers modest gains
($+0.009$ to $+0.035$ NSE). On Methane---where the flux depends on soil
and hydrological factors absent from the exogenous inputs
(cf.\ \S\,\ref{sec:prior_robustness})---the regression signal is weaker,
the prior is underdetermined, and co-adaptation degrades zero-shot
transfer ($-0.030$). This is consistent with Methane being the dataset on
which cross-location transfer is most fragile
(Table~\ref{tab:main_results}).

Two-stage training enforces gradient isolation between the prior and the
denoiser, avoiding co-adaptation by construction---a robust default for
zero-shot reconstruction. Joint optimization is a viable refinement when
domain knowledge indicates a strong, well-identified
$\mathbf{X}\to\mathbf{Y}$ coupling. Whether joint schemes can retain
these gains while suppressing co-adaptation remains an open direction.

\clearpage
\section{Hyperparameter Search}
\label{appendix:hyper}

\begin{figure}[H]
    \centering
    \includegraphics[width=\textwidth]{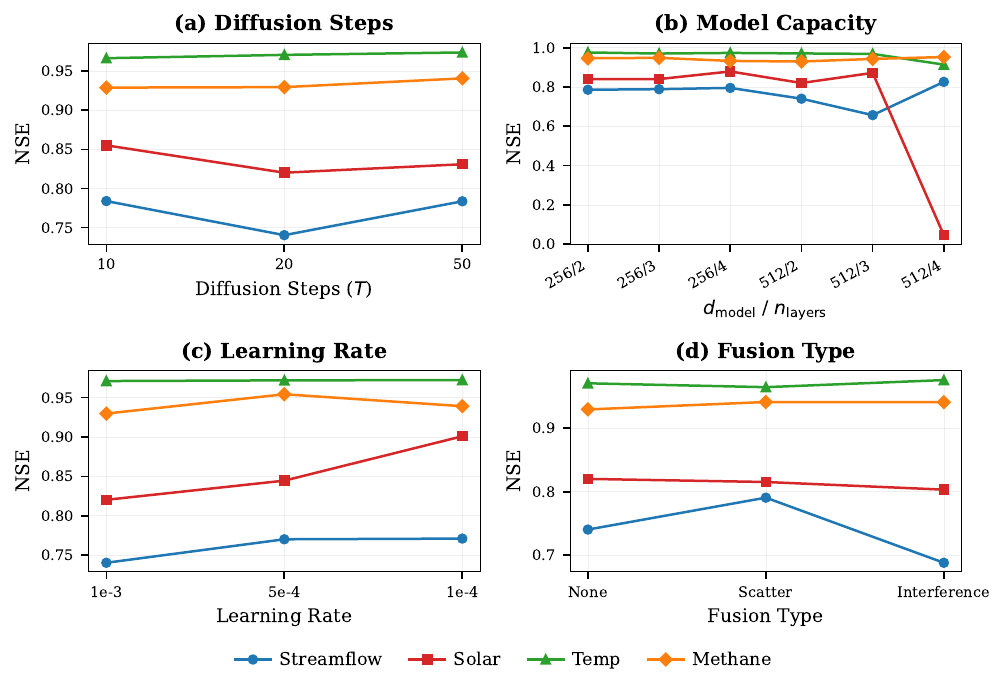}
    \caption{Hyperparameter sensitivity of ZeroDiff across four datasets, measured by NSE. Each panel varies one factor: (a) diffusion steps $T$, (b) denoiser capacity ($d_{\mathrm{model}}$ / $n_{\mathrm{layers}}$), (c) learning rate, and (d) exogenous fusion strategy.}
    \label{fig:hyper}
\end{figure}

We conduct a systematic hyperparameter sensitivity analysis for the diffusion component of ZeroDiff across all four datasets. We vary one factor at a time while keeping the remaining settings at their defaults ($d_{\mathrm{model}}=512$, $n_{\mathrm{layers}}=2$, $T=20$, $\mathrm{lr}=10^{-3}$, no fusion). Results are summarized in Figure~\ref{fig:hyper}.

\paragraph{Diffusion steps.}
Increasing $T$ from 10 to 50 yields modest but consistent gains on most datasets, with $T=50$ achieving the best NSE on Streamflow, Temp, and Methane. The improvement is marginal relative to the added computational cost, so we adopt $T=20$ as the default for efficiency.

\paragraph{Model capacity.}
Smaller denoisers ($d_{\mathrm{model}}=256$) perform comparably to or better than larger ones ($d_{\mathrm{model}}=512$) at moderate depth. However, over-parameterization can be harmful: the $512/4$ configuration causes a severe performance collapse on Solar (NSE $= 0.05$) and a notable drop on Temp, indicating overfitting when the denoiser capacity exceeds what the calibration task requires.

\paragraph{Learning rate.}
Performance is relatively stable across the tested range ($10^{-3}$ to $10^{-4}$). Solar benefits most from a lower learning rate ($10^{-4}$), while other datasets show mild improvements or remain flat. We use $10^{-3}$ as the default.

\paragraph{Fusion type.}
The exogenous fusion strategy has a dataset-dependent effect. Scatter fusion improves Streamflow and Methane, while interference fusion benefits Temp. No single fusion strategy dominates across all datasets, so we default to no fusion for simplicity and leave fusion selection as a dataset-specific tuning option.